\pdfoutput=1

\documentclass[journal]{IEEEtran}
\usepackage{times,latexsym}
\usepackage{url}

\usepackage{amsmath}
\usepackage{amssymb}
\usepackage{caption}
\usepackage{subcaption}
\usepackage{enumerate}
\usepackage{multirow}
\usepackage{graphicx} 
\usepackage{url}
\usepackage{color}
\usepackage{booktabs}
\usepackage{multicol}

\usepackage{cleveref}
\crefname{section}{§}{§§}

\usepackage{acronym}

\usepackage[numbers]{natbib}

\acrodef{MRC}{machine reading comprehension}
\acrodef{RC}{reading comprehension}

\ifCLASSINFOpdf
\else
\fi
\begin{document}
%
\title{From LSAT: The Progress and Challenges of Complex Reasoning}
%
%
%

\author{
        Siyuan Wang,~\IEEEmembership{Student Member, IEEE,}
        Zhongkun Liu, Wanjun Zhong, 
        Ming Zhou,~\IEEEmembership{Member, IEEE,} \\
        Zhongyu Wei,~\IEEEmembership{Member, IEEE,}
        Zhumin Chen,
        and~Nan Duan,~\IEEEmembership{Member, IEEE,}

\thanks{Work is done during internship at Lanboat. Corresponding author: Ming Zhou; Zhongyu Wei.}
\thanks{Siyuan Wang and Zhongyu Wei are with the School of Data Science, Fudan University, Shanghai 200433, China. And Zhongyu Wei is also with the Research Institute of Intelligent and Complex Systems, Fudan University, Shanghai 200433, China
(email: wangsy18@fudan.edu.cn; zywei@fudan.edu.cn).}
\thanks{Zhongkun Liu and Zhumin Chen are with the School of Computer Science and Technology, Shandong University, Qingdao 266237, China (email: liuzhongkun@mail.sdu.edu.cn; chenzhumin@mail.sdu.edu.cn)}
\thanks{Wanjun Zhong is with the School of Computer Science and Engineering, Sun Yat-Sen University, Guangzhou 510275, China (email: zhongwj25@mail2.sysu.edu.cn).}
\thanks{Ming Zhou is with the Sinovation Ventures, Beijing 100080, China (email: zhouming@chuangxin.com).}
\thanks{Nan Duan is with Microsoft Research, Beijing 100080, China (email: nanduan@microsoft.com).}
}

%
%

\ifCLASSOPTIONpeerreview
\markboth{IEEE/ACM TRANSACTIONS ON AUDIO, SPEECH, AND LANGUAGE PROCESSING, 2021}%
{\MakeLowercase{\textit{et al.}}: From LSAT: The Progress and Challenges ofComplex Reasoning}
\fi
%



\maketitle

\begin{abstract}
Complex reasoning aims to draw a correct inference based on complex rules. As a hallmark of human intelligence, it involves a degree of explicit reading comprehension, interpretation of logical knowledge and complex rule application. In this paper, we take a step forward in complex reasoning by systematically studying the three challenging and domain-general tasks of the Law School Admission Test (LSAT), including analytical reasoning, logical reasoning and reading comprehension.  
We propose a hybrid reasoning system to integrate these three tasks and achieve impressive overall performance on the LSAT tests. The experimental results demonstrate that our system endows itself a certain complex reasoning ability, especially the fundamental reading comprehension and challenging logical reasoning capacities. Further analysis also shows the effectiveness of combining the pre-trained models with the task-specific reasoning module, and integrating symbolic knowledge into discrete interpretable reasoning steps in complex reasoning. We further shed a light on the potential future directions, like unsupervised symbolic knowledge extraction, model interpretability, few-shot learning and comprehensive benchmark for complex reasoning.
\end{abstract}

\begin{IEEEkeywords}
LSAT, complex reasoning, analytical reasoning, logical reasoning, reading comprehension.
\end{IEEEkeywords}

%
\IEEEpeerreviewmaketitle

%
%
%
%

\section{Introduction}
\IEEEPARstart{C}{omplex} reasoning aims to comprehend and analyze the given information, and apply complex rules to draw correct inference~\cite{duschl2007taking, songer2009and}.
As an essential ability for complex problem solving,
it provides tremendous opportunities for
many real-world scenarios, such as mathematical word problems, negotiation and argument, and medical diagnosis~\cite{chattopadhyay2013case, shi2015automatically, reinhold2020role, galassi2020neural}.
In recent years, having a computer pass admission examinations is a hot AI challenge towards complex reasoning, which offers an objective and accurate measurement with a certain difficulty.
\citet{fujita2014overview} design a system to sit the Japanese National Center Test for University Admissions. Gaokao, as the National College Entrance Examination of China, also has been widely studied~\cite{cheng2016taking,yu2016history}. Although encouraging results have been achieved in taking these real high-school exams,
these works study domain-specific and limited complex reasoning capabilities.  
They cover different subjects and each is separately processed with subject-specific knowledge~\cite{zhang2018towards}. For example, the annotated formulas representing physics, mathematics and biology knowledge are usually assumed to be provided to solve corresponding problems~\cite{fujita2014overview, gunning2010project}, while history and geography questions are primarily solved by retrieving relevant information supplemented by shallow reasoning to match the answer~\cite{ding2018answering,zhang2018one}.


\begin{figure*}[!th]
\centering
\includegraphics[width=1.95\columnwidth]{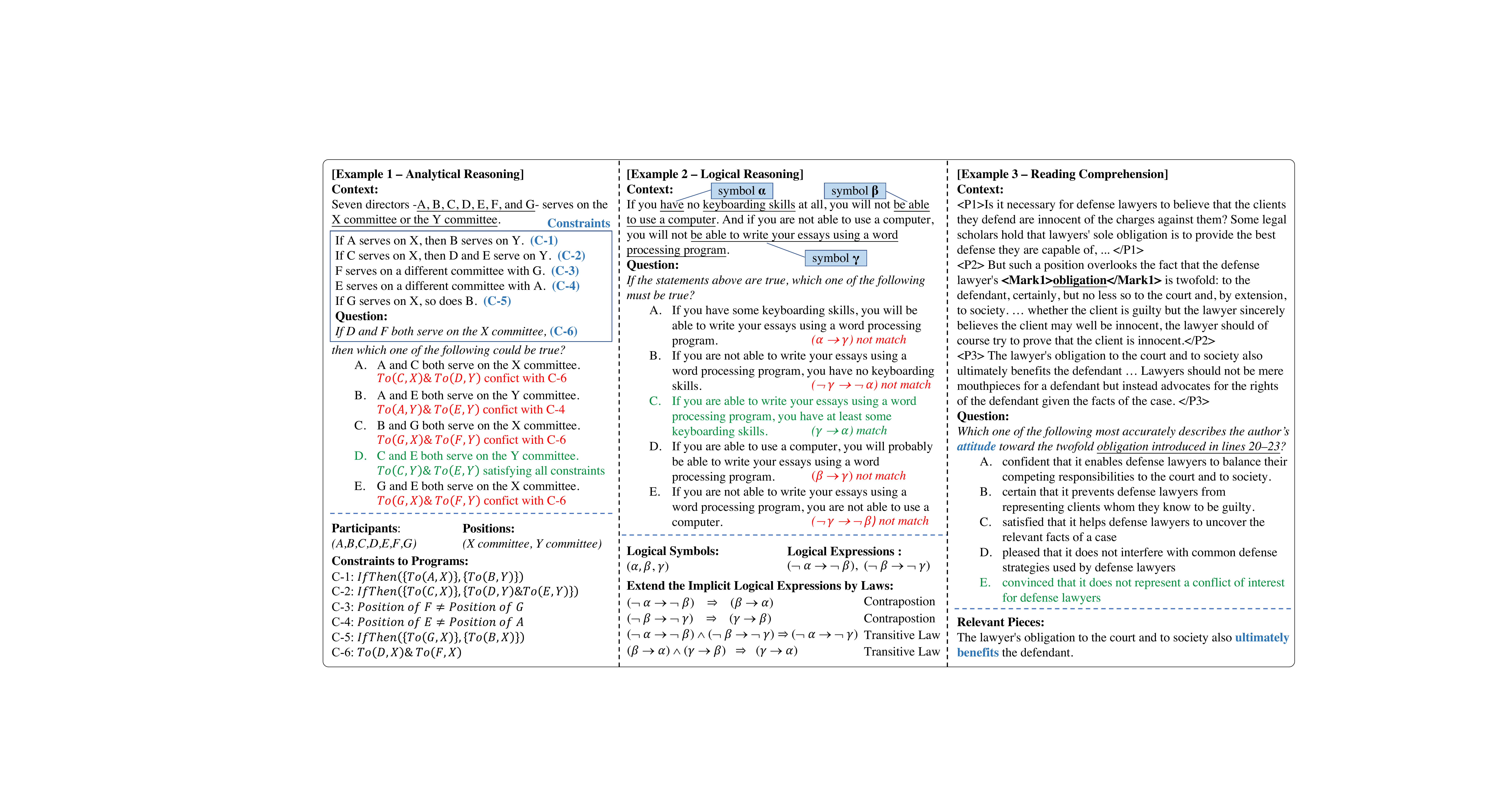}
\caption{\label{figure_example} Three examples of LSAT tasks with the required reasoning processes. For AR, it needs to understand the knowledge of participants, positions, and constraints, and deduce the legitimate option. For LR, the elementary logical symbols and expressions need to be identified to logically infer the implicit expression. For RC, it requires locating the relevant pieces by the positional indicators and abstract the answer.
The options in \textcolor[RGB]{74,163,87}{green} mean the correct answers of three examples.}
\end{figure*}
To take a step towards a more challenging and domain-general complex reasoning ability, we focus on the Law School Admission Test (LSAT)\footnote{\url{https://www:lsac:org/lsat/}.},
which is one of the most difficult exams covering multiple domains.
LSAT is a standardized test administered for prospective law school candidates worldwide, which mainly assesses their general complex reasoning skills, including analytical reasoning, logical reasoning, and reading comprehension capabilities in general domains.
Correspondingly, the LSAT can be categorized into three tasks:
(1) analytical reasoning (AR), measures the ability to analyze a scenario ruled by a set of constraints, and determine which option satisfies or conflicts with all the constraints. (2) logical reasoning (LR), a task that focuses on the logical analysis of texts and performing logical inference to deduce implications from asserted ones.
(3) \ac{RC}, revolves around the ability to deeply understand long-form materials, and locate the relevant pieces to distinguishing what is the case by summarizing or comparing highly abstract concepts, such as attitudes and principles.
Three examples of LSAT tasks are listed in Figure~\ref{figure_example}, which are all quite challenging and involve complex reasoning processes.
Therefore, we aim to systematically explore the progress and challenges of complex reasoning, and take the real-world LSAT tasks which have been rarely explored~\cite{yu2020reclor} as a testbed.  

Existing methods for complex reasoning can be summarized into three types, namely, symbolic models, neural models, and neural-symbolic models~\cite{itzhaky2013solving, bach2015hinge}. 
Symbolic models identify the discrete symbols (like entities and logical functions) as basic reasoning units, 
and perform explicit inferences upon symbolic representations. With controllable and interpretable reasoning steps, yet they largely depend on expert-defined rules which are inflexible for different datasets and lack resilience against data noises~\cite{zhang2021neural}. 
Neural models mimic the neuron connections in the human brain to learn the semantics of data with continuous vectors and implicitly infer the answer, which are robust to the ambiguous and noisy data but short of interpretability. To achieve synergies among the advantages and circumvent the limitations of symbolism and connectionism, neural-symbolic models integrate both symbolic logic and continuous representation to reason out the answer~\cite{chen2019neural}. 

From the above perspectives, we design a hybrid reasoning system for three LSAT tasks.
For AR, we first propose a symbolic system~\cite{zhong2021arlsat} which designs rules to identify symbolic participants and constraints, and deterministically deduce the legitimate solutions. We then attempt a neural method utilizing a graph network for modeling constraints between participants.
We also come up with a neural-symbolic model which neurally parses the textual constraints into programs and discretely executes the programs to reach the answer.
For LR, we propose a neural-symbolic logic-driven system~\cite{wang2021logicdriven}, which employs a symbolic module to extract logic from the texts and infer the entailed logic by logical laws, then utilizes a neural module to encode the inferred logic for answer prediction.
For \ac{RC}, we propose a neural Transformer-based approach with an external multi-head attention mechanism~\cite{zhu2020duma} and passage positional information for better understanding the interaction between context and question.

\begin{figure*}[!th]
\centering
\includegraphics[width=1.72\columnwidth]{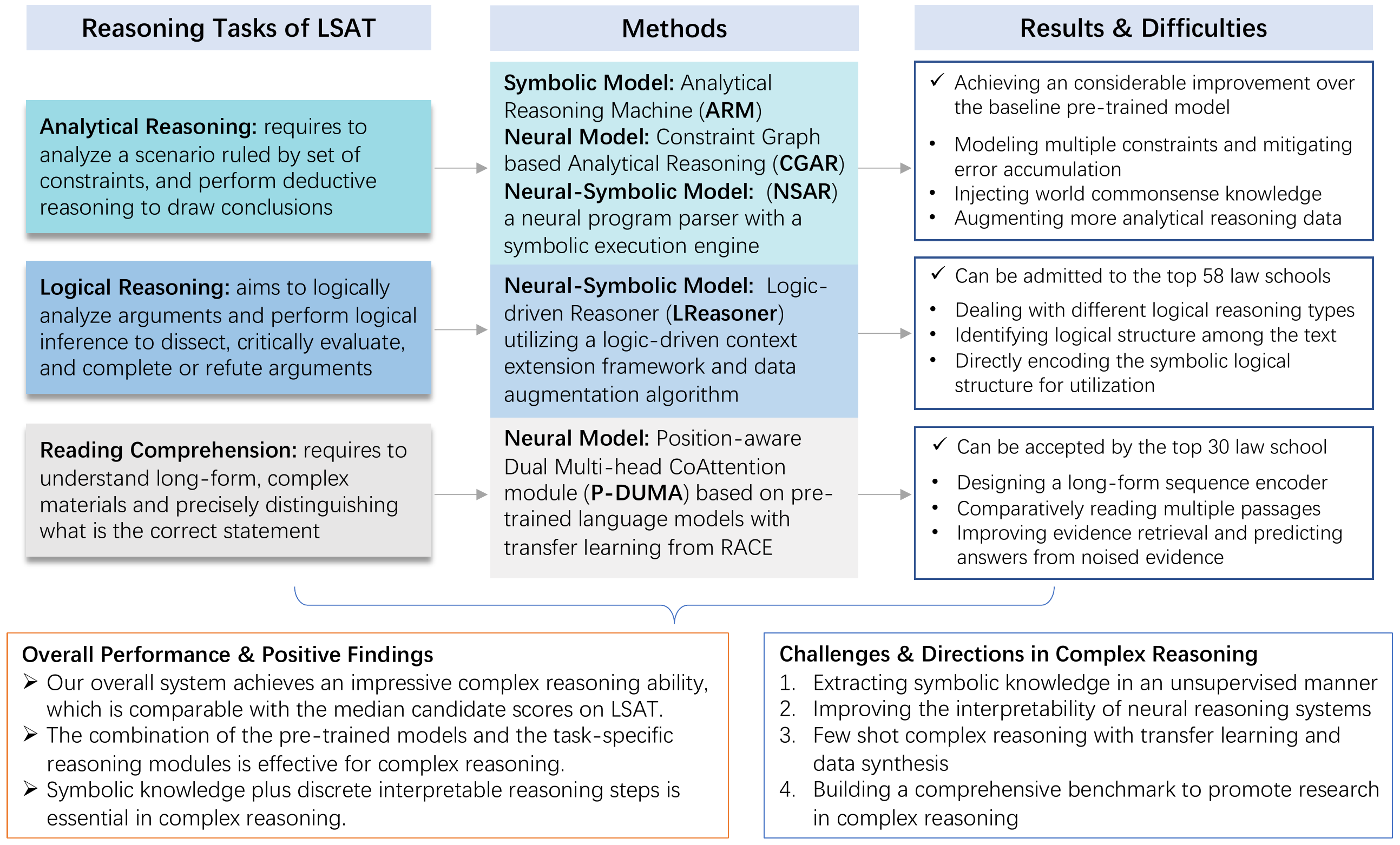}
\caption{\label{figure_illustration} The overall illustration of our investigation from LSAT towards complex reasoning.}
\end{figure*}

On the whole, our overall system integrating three tasks achieves an accuracy of $56.8\%$ on the LSAT exams, which is comparable to the 
median for human test taker scores.
Our systems for RC and LR even have a chance to be admitted to the top 30 and top 58 law schools, respectively. The results show the effectiveness of our systems for modeling complex reasoning abilities, notably the fundamental reading comprehension and challenging logical reasoning capability.
The overall illustration of our investigation on LSAT towards complex reasoning is shown in Figure~\ref{figure_illustration}. Through a systematical study of three LSAT tasks, we not only achieve great performance on LSAT, but also make some progress towards complex reasoning. We further investigate the emerging challenges and inspire potential future directions of complex reasoning. For example, automatically extracting symbolic knowledge in an unsupervised manner, few-shot complex reasoning, improving the interpretability of the neural reasoning system, and building a comprehensive benchmark, are all essential to be explored to promote complex reasoning research. 

In the rest of the paper, we first list some related work of complex reasoning and corresponding advanced methods. We also preliminarily introduce the tasks, baseline models and datasets of LSAT in \cref{related_work}. Then we separately introduce the challenges, detailed methods, experimental results and further analysis for analytical reasoning, logical reasoning and reading comprehension tasks in \cref{analytical_reasoning}, \cref{logical_reasoning} and \cref{reading_comprehension}, respectively. We further summarize our overall performance on LSAT, and discuss the major challenges and future directions in complex reasoning in \cref{overall_discussion}. We finally draw our conclusion in \cref{conclusion}.

\section{Related Work}
\label{related_work}
\subsection{Taxonomy of Complex Reasoning}
\label{reasoning_taxonomy}
To encourage the progress of artificial intelligence systems towards deeper human-like comprehension and reasoning, there has been a surge in complex reasoning research in recent years. We first investigate  existing works on several major aspects of complex reasoning.

\paragraph{Logical Reasoning} An increasing number of tasks and datasets have been introduced targeting logical reasoning. Natural Language Inference \cite{dagan2005pascal, bowman2015large, williams2017broad} aims to determine the entailment relationship between a hypothesis and a premise, which requires relatively simple logical reasoning ability at the sentence level.
Several question answering datasets have been proposed for promoting logical reasoning ability, i.e., LogiQA \cite{liu2020logiqa}, ReClor~\cite{yu2020reclor}, which are sourced from public standardized exams. However, previous methods usually fail to model the discrete logical inference process explicitly.
This work also dives into logical reasoning, and considers understanding the elementary logical structure and perform explicit logical inference to draw a logical conclusion.
\paragraph{Commonsense Reasoning} Commonsense reasoning requires utilizing commonsense knowledge to reason out the answer, which attracts great concern of the research communities. Recently many benchmarks have been introduced to assess reasoning capabilities over different commonsense knowledge, such as domain-specific knowledge~\cite{rashkin2018event2mind,talmor2018commonsenseqa,zhou2019going}, general semantic knowledge~\cite{speer2017conceptnet} and inferential knowledge~\cite{sap2019atomic}. Current methods, however, are still not robust enough to be deployed in the open domain and ignore directly modeling commonsense through symbolic integration~\cite{sap2020introductory}.
How to incorporate symbolic commonsense and reason over relevant knowledge will be substantially explored~\cite{zhang2018record, huang2019cosmos}. 
\paragraph{Multi-Hop Reasoning} Multi-hop reasoning is another widely studied complex reasoning task recently~\cite{khashabi2018looking, welbl2018constructing, yang2018hotpotqa, inoue2019r4c}, which requires reasoning across multiple pieces of sentences or documents to model multi-hop relationships to reach the answer. Conversational question answering tasks~\cite{choi2018quac, reddy2019coqa} also involve multi-hop reasoning over multi-turn utterances. Constructing correct multi-hop reasoning chains (conversation flows) has been a key
challenge for these tasks, and it is an essential research direction to effectively model multi-hop reasoning paths over multiple passages\cite{chen2019multi, wang2019multi}.
\paragraph{Numerical Reasoning} Numerical reasoning involves performing discrete arithmetic reasoning over quantities to solve mathematical word problems\cite{clark2016combining, ling2017program, dua2019drop, amini2019mathqa}, which is a fundamental and challenging task. Previous work translates the textual math word problem into an expression or an expression tree and utilizes arithmetic knowledge to solve it~\cite{ling2017program, wang2018translating, xie2019goal}. With weak supervision, how to improve the accuracy of generated expression trees is worth further study~\cite{hong2021learning, saha2021weakly}. 

\subsection{Advanced Methods of Complex Reasoning}
Advanced methods to solve complex reasoning problems, no matter what specific reasoning skill is required, can be summarized as following three types: symbolic models, neural models and neural-symbolic models~\cite{itzhaky2013solving, bach2015hinge, zhang2021neural}.

\paragraph{Symbolic Models} 
As complex reasoning requires discrete reasoning operations over reasoning units, most previous studies for complex reasoning are symbolic expert systems~\cite{galarraga2013amie, zheng2015build, minsky2017perceptrons, donadello2017logic}. They design a set of rules or templates to identify basic units for reasoning as symbols, such as the quantities and arithmetic signs for numerical reasoning~\cite{fuchs2013effects, supekar2013neural, donadello2016integration}, relevant triples and paths from knowledge graph for commonsense reasoning~\cite{sun1995robust}, etc. They then perform deterministic and explicit inferences upon discrete elementary units to predict the answer. The symbolic models provide sound readability and interpretability. 
However, it requires expert knowledge and tremendous human efforts in designing rules, which makes it inflexible to scale across different datasets and lack resilience against data noises.
Moreover, the finite and discrete symbolic representations are insufficient to depict all the intrinsic reasoning structures.
\paragraph{Neural Models}
Neural models mimic the neuron connections in the human brain~\cite{rosenblatt1958perceptron, rumelhart1986learning} and apply neural networks to implicitly represent the abstracted semantics of input text and knowledge with continuous vectors, which are robust to the ambiguous and noisy data
~\cite{hinton2006fast, bengio2007greedy}. For example, pre-trained language models have achieved superior performance on many comprehensive tasks~\cite{devlin2019bert, liu2019roberta, lan2019albert}. However, the decision process of neural models is always a black box, which makes the prediction lacks explainability and reliability.
Whether the neural models show the reasoning ability or just capture the data bias to achieve high accuracy is also a question. Graph neural networks~\cite{kipf2016semi, velivckovic2017graph} and neural module networks~\cite{andreas2016neural, gupta2019neural} are also introduced to partly imitate the human reasoning process to make up interpretability. However, these methods still perform an implicit inference to reason out the answer without a clue as to why and how. 
Besides, neural models depend a lot on training data and are computationally expensive to train, and the performance will sharply decrease with limited data and resource. 

\paragraph{Neural-Symbolic Models} To combine the advantages and circumvent the shortcomings of both symbolic and neural methods, neural-symbolic models which integrate symbolic logic and neural representation are widely studied~\cite{gallant1988connectionist, gallant1993neural, de2011neural, besold2017neural, garcez2019neural}. 
Some work employs a neural module to parse the language into executable programs~\cite{liang2016neural, chen2019neural} and deterministically executing the programs to find the answer in a symbolic module. For example, the neural-symbolic models for numerical reasoning translate the input texts into expressions through a neural generation model and then discretely execute the expressions~\cite{ling2017program, wang2018translating, xie2019goal}. Other work first designs rules to extract the reasoning units and explicitly conduct inference over them in a symbolic module. They then utilize a neural module to learn continuous vectors for symbolic representations to deal with the uncertainty of data~\cite{zhong2020logicalfactchecker, huang2021dagn, wang2021logicdriven}. 
Specifically, for commonsense and multi-hop reasoning, the symbolic module can be designed to identify relevant knowledge and multi-hop reasoning chain, respectively, and then encode them into the neural module to match the answer~\cite{arabshahi2020conversational, liu2020integrating, liu2021neural, moghimifar2021neural}. However, how to generate high-quality programs under weak supervision and extract symbolic reasoning units in an unsupervised manner remains an elusive challenge.

\subsection{Examination-based Question Answering}
Recent years have witnessed an emerging trend in answering complex questions collected from human standardized examinations at a different level of education, which is more difficult and measurable.
The Todai Robot project aims to create a system that can pass the Japanese National Center Test for University Admissions~\cite{fujita2014overview}. Aristo Challenge focused on solving the questions of Elementary School Science and Math Tests which is for 6-11 year olds~\cite{clark2015elementary}. A similar project studying the National College Entrance Examination of China (Gaokao) also has been launched~\cite{cheng2016taking,yu2016history,ding2018answering}. These challenges deal with various subjects, like mathematics, biology, physics, geography, history, and drive the progress of AI systems towards complex problem solving and reasoning. However, these questions rely heavily on domain-specific expressions and knowledge like formulas in mathematics and physics, and quotes in Classical Chinese, which ignore the domain-general complex reasoning ability. 
Besides, the studies of these tasks have hit the bottleneck of commonsense-based reading comprehension and general intelligence, which fail to be admitted to key universities~\cite{aivs2018ai}.

RACE dataset~\cite{lai2017race} is introduced to remove domain restrictions by collecting the general English exams for middle and high school Chinese students. However, around 70\% of questions are in the category of word matching, paraphrasing or single-sentence reasoning, which are relatively simple.
LogiQA \cite{liu2020logiqa} collected from the National Civil Servants Examination of China and ReClor \cite{yu2020reclor} from the Graduate Management Admission Tests and Law School Admission Tests both require deeper logical reasoning. In this paper, we not only examine the logical reasoning capability of LSAT, but also challenging analytical reasoning and complicated reading comprehension abilities to systematically explore complex reasoning.

\section{Preliminaries}
\label{preliminaries}
\subsection{Task Definition}
The LSAT problems are formulated as a multiple-choice question answering task, which is described as follows. Given a context $c$, a question $q$ together with five candidate options $o={o_1, o_2, o_3, o_4, o_5}$, only one option is need to be predicted as the most plausible answer $o_a$.

\subsection{Baseline Model}
\label{sec:base_model}
Pre-trained Transformer-based language models, i.e., BERT \cite{devlin2019bert}, RoBERTa~\cite{liu2019roberta} and ALBERT~\cite{lan2019albert}, achieve impressive performance on multiple-choice question answering, which can be employed as the baseline model of all LSAT tasks.
Specifically, the context, the question and an option are concatenated as the input for encoding, which is formulated as $[CLS] \ c \ [SEP] \ q \ || \ o_i \ [SEP]$ and $||$ is the concatenation operator. 
Given five options, five concatenated sequences are constructed to be encoded. The representations of special token $[CLS]$ in five sequences are fed into a classification layer to get the probabilities of options as their scores, and the option with the highest score is selected as the answer. The models are fine-tuned with cross-entropy loss on the training set.  


\subsection{Dataset}
The LSAT datasets are collected from previous exams from 1991 to 2016,
including a total of 90 examinations. Each exam roughly contains 100 questions, among which 1/2 are logical reasoning questions, 1/4 are analytical reasoning questions and the rest 1/4 are reading comprehension questions. For a small proportion of questions with only four options, we randomly select one of the wrong choices as a supplemental option for each instance. We name datasets of analytical reasoning, logical reasoning and reading comprehension tasks as AR-LSAT~\cite{zhong2021arlsat}, LR-LSAT and RC-LSAT, respectively. Each dataset is further split into training, validation and testing sets. 
The detailed statistics of each dataset are listed in Table~\ref{table_statistics}. 
\begin{table}[!h]
\begin{center}
\resizebox{0.485\textwidth}{!}{
\setlength\tabcolsep{4.5pt}
\begin{tabular}{c|c|cccc}
\toprule
Datasets & Statistics & Train & Val & Test & Total \\
\toprule
\multirow{2}{0.6in}{\emph{AR-LSAT}} & \emph{\# context} & 280 & 40 & 40 & 360 \\
 & \emph{\# question} & 1,630 & 231 & 230 & 2,091 \\
\midrule
\multirow{2}{0.6in}{\emph{LR-LSAT}} & \emph{\# context} &3,325 & 503 & 507 & 4,335 \\
 & \emph{\# question} & 3,529 & 506 & 510 & 4,545 \\
\midrule
\multirow{2}{0.6in}{\emph{RC-LSAT}} & \emph{\# context} & 280 & 40 & 40 & 360 \\
 & \emph{\# question} & 1,880 & 270 & 269 & 2,419 \\
\bottomrule
\end{tabular}
}
\caption{The detailed statistics of LSAT datasets. \emph{\# context} and \emph{\# question} are the numbers of contexts and questions in each split.}
\label{table_statistics}
\end{center}
\end{table}



As collected from standardized examinations, these LSAT datasets are of high quality and difficulty for complex reasoning, and are accompanied by an accurate evaluation metric and relatively limited data size. The data sparsity makes these reasoning tasks more difficult to be solved by traditional data-driven approaches that purely learn data patterns from massive data. Therefore, the models with solid complex reasoning abilities need to be developed and 
decrease the dependency on the data size.


\section{Analytical Reasoning}
\label{analytical_reasoning}
\subsection{Challenges in Analytical Reasoning} 
\label{ar_challenge}
Analytical Reasoning aims to analyze a scenario involving a set of predefined constraints and perform deductive reasoning to draw correct solutions. As no previous work or benchmark dataset completely studies this challenging task, we introduce a new dataset, namely AR-LSAT~\cite{zhong2021arlsat}, to foster research on this area. Most of the questions in AR-LSAT can be viewed as a constraint satisfaction problem~\cite{kumar1992algorithms}, which needs to find legal assignments of positions to participants satisfying the given constraints.
All participants, positions and constraints are described in the context. Take a look at the first example in Figure~\ref{figure_example}, two positions (i.e., X committee and Y committee) need to be assigned to seven directors (i.e., A, B, etc.) under a set of constraints. To solve such problems, the system requires understanding the game settings including the compositions of participants, the possible values of positions, and interpreting the logical meaning of the constraints descriptions. Then it involves conducting inference over constraints to deduce the answers.

The analytical reasoning problems are quite challenging.
The situation descriptions in the context are diverse with no domain restriction. 
A more accurate and comprehensive understanding of the context is also required, because each piece of the context is significant in building the whole reasoning chain.
Besides, the five options of a question tend to be similar to each other and the answer never explicitly appears in the context. 
Therefore, the AR task cannot be superficially solved by relevant information extraction and shallow contextual matching.

\subsection{Symbolic Model: ARM}
\label{sec:rule_model}
To explicitly model discrete analytical reasoning steps, we start with a symbolic model called Analytical Reasoning Machine (ARM)~\cite{zhong2021arlsat}, which can answer the question by pre-defined rules and deterministic deduction.
To solve questions like the first example in Figure\ref{figure_example}, we propose to perform a four-stage reasoning process:
1) extracting participants, positions, constraints from the context and question;
2) interpreting constraints into executable programs, i.e., a combination of logical functions;
3) generating a set of legitimate assignments by executing all programs;
4) selecting the most plausible option by matching the legitimate assignments.
Next, we will introduce these steps in detail.

\subsubsection{Arguments Extraction}
We first extract participants, positions and constraints from the context to have a primary understanding of the problem. 
Specifically, we use a Named Entity Recognition model~\cite{peters2017semi} to extract entities from the context and group them into participants and positions. Entities that appear together in the leading sentence of the context are recognized as participants or positions, where participants always appear before positions.
We also identify constraints by judging whether a sentence restricts the assignments between participants and positions. As the example in Figure~\ref{figure_example}, seven participants (i.e., \emph{A,B,C,D,E,F,G}) need to be assigned into two positions (i.e., \emph{X committee and Y committee}). All sentences except the first are recognized as constraint descriptions.

\subsubsection{Constraints Interpretation}
\label{constraints_interpretation}
We next interpret natural language constraints into executable programs based on three types of predefined functions, i.e., relational, counting and compositional functions.
The relational function involves participants and positions as arguments and focuses on the relationship between them, e.g., ``\emph{Before(A,B)}'' means ``\emph{A must be in a lower-numbered position than B}'' while ``\emph{To(A,X)}'' indicates ``\emph{Participant A is assigned to position X}''. The counting function describes the numerical and order constraints over participants, which take as arguments both participants and numbers.
A compositional function takes two sets of relational or counting functions as arguments and formulates the relationship between them, like conditional (\emph{if-then}) relationship. For example, the constraint ``\emph{If A serves on the X, then B serves on the Y}'' can be expressed as ``\emph{IfThen(\{To(A,X)\}, \{To(B,Y)\})}''. 

We design a set of trigger words to match potential functions~\cite{liang2016neural} and extract arguments~(i.e., participants, positions, and numbers) according to their relative positions to the trigger words.
For uncertain sentences with no matched function, we also build a neural classification model based on RoBERTa to predict their corresponding function types. The combination of functions in each sentence is the interpreted program. In figure~\ref{figure_example}, the set of interpreted programs corresponding to the AR example is also presented.

\begin{figure}[ht]
    \centering
    \includegraphics[width=0.95\columnwidth]{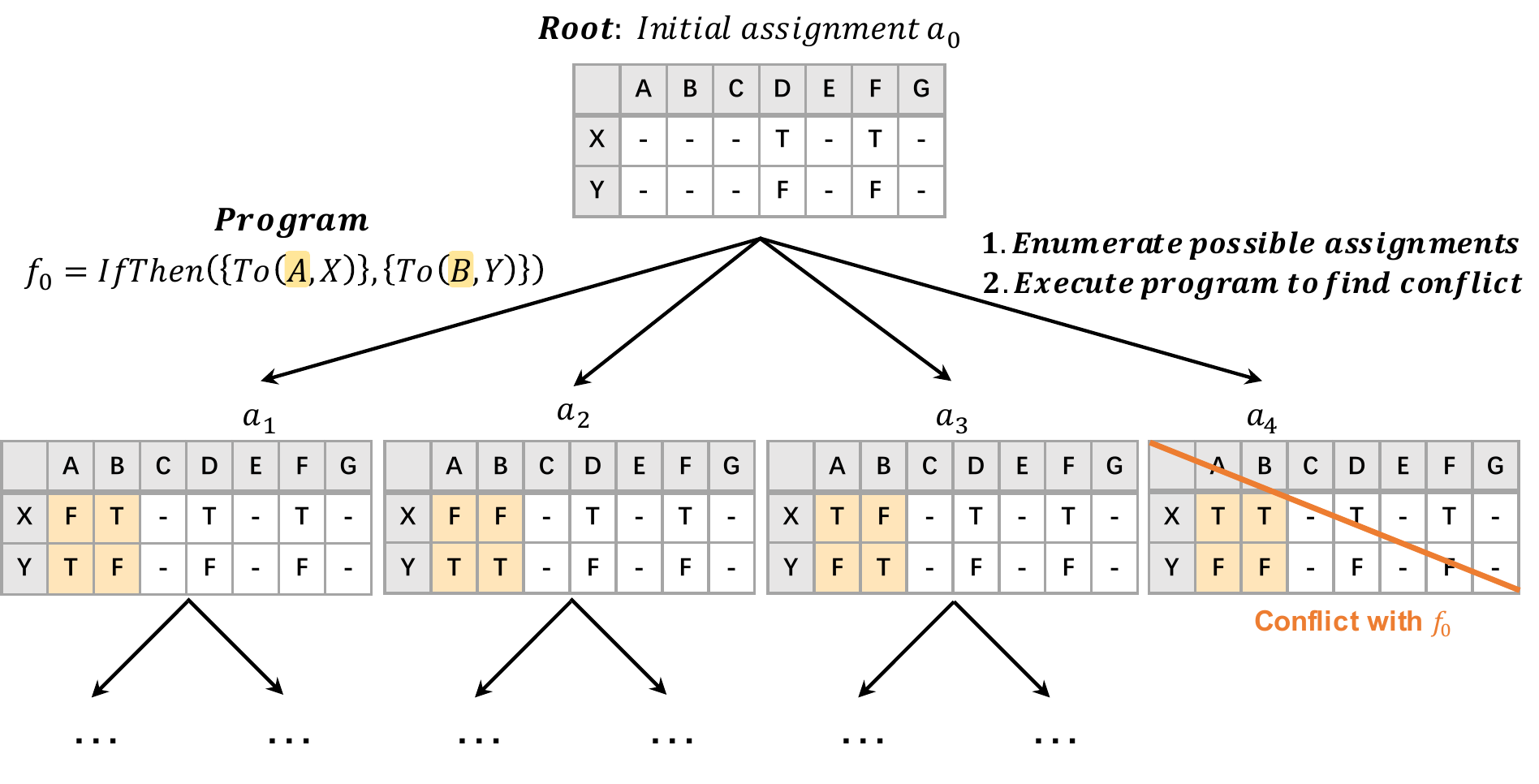}
    \caption{The tree-based reasoning process. Here ``T/F/-'' means ``True/False/Unknown''.}
    \label{fig_execution}
\end{figure}

\subsubsection{Program Execution}
\label{Program_exe}
We recognize the programs that immediately determine the positions of some participants to construct an initial assignment. For example, the constraint ``\emph{D and F both serve on the X committee}'' corresponds to the initial assignment in Figure~\ref{fig_execution}. Each assignment is formulated as a table with columns as participants and rows as positions.
Each cell in the assignment has three possible states, i.e., \emph{True, False, Unknown}, which means whether a participant is assigned to a position.

Starting from the initial assignment, we conduct reasoning to deduce legitimate assignments satisfying all constraints by executing other programs. The reasoning process is formulated into a tree-based heuristic shown in Figure~\ref{fig_execution}. Specifically, each node is an assignment while each edge is an executable program, and the initial assignment is taken as the root node $a_0$.
At each iteration, for a program $f_i$ we enumerate all possible positions for participants involved in this program if they are unknown,
and execute the program to check satisfaction for each assignment to remove the illegitimate ones.
Then taking each legitimate assignment as a new root and starting a new iteration, we validate the next program to further extend the reasoning tree. The tree is recursively expanded until all programs are executed. The leaf nodes that satisfy all constraints are obtained as final legitimate assignments.

\subsubsection{Option Selection}
\label{option_select}

After deducing all legitimate assignments, we analyze the options to select the most plausible one as the answer.
As each option can be interpreted as a program in the same way described in \cref{constraints_interpretation}, we can further extend the reasoning tree and execute the option-based program.
We take the number of legitimate assignments of each option as its score and the option with the maximum score is selected as the final answer. 

\subsection{Neural Model: CGAR}
As ARM requires substantial effort to craft sets of rules which are far from perfect, we attempt to design two neural models to ease the manual work. We first adopt the state-of-the-art pre-trained models including RoBERTa~\cite{liu2019roberta} and ALBERT~\cite{lan2019albert} as described in \cref{sec:base_model}. However, they struggle to capture complex reasoning ability beyond shallow-level semantic understanding and perform nearly a random guess~\cite{zhong2021arlsat}.

We also propose a Constraint Graph-based Analytical Reasoning (CGAR) framework. Considering that for each AR problem, participants are to be assigned into several positions, and each constraint describes a restriction relationship between some participants and positions. To better model the relationship structure of involved participants and positions to deduce the legitimate solutions, we propose constructing constraint graphs for each question. We thus introduce a CGAR framework that utilizes a graph convolutional network (GCN)~\cite{kipf2017semi} and a pre-trained model to reason over the constraint structure. The framework is composed of three modules, namely, a graph construction module, a graph reasoning module and an answer prediction module.

\subsubsection{Constraint Graph Construction}
We construct a constraint graph $\mathcal{G}_i$ for each (context $c$, question $q$, option $o_i$) triple. Then five graphs can be constructed corresponding to five options of a question. Each is designed as a heterogeneous undirected graph where the nodes consist of participants, positions and constraints in the context. Each constraint node is connected to its mentioned participant nodes and position nodes.
Besides, the (question, option) pair is also injected as a global node to establish linkages with all constraint nodes, to indicate whether the option satisfies the constraints and answers the question.
We follow the extraction method in \cite{zhong2021arlsat} which utilizes named entity recognition to extract participants and positions from the leading sentence of the context. And we simply take each sentence in the context as a constraint.

\subsubsection{Graph Reasoning}
To initialize the node representations, we utilize the output of the pre-trained model as the embedding of each token. For participant and position nodes, we perform mean pooling on the constituent token embeddings to get their representations. For constraint nodes, we take the average of the start and end token embeddings of each sentence as their representations. 
Then two $[SEP]$ token embeddings are also averaged to initialize the representation of the global node. 
To model the heterogeneity of the graph, we also define three types of nodes, including the entity node, the constraint node and the global node. Participant and position nodes are both viewed as entity nodes. We utilize a linear transformation onto the node representations for different node types to get their type-specific representations. 

We employ a GCN to perform reasoning over a constraint graph $\mathcal{G}_i$. During each message-passing iteration, we hope to model the feasible states of participants and positions by aggregating the constraint information to entities. Correspondingly, constraint features and global features are also updated. After multiple iterations, the global node is aware of available states of participants and positions in (question, option) pair for predicting whether option $o_i$ is the correct answer.

\subsubsection{Answer Prediction}
As in \cref{sec:base_model}, we also need to compute scores of each option to find the most plausible answer. We determine the plausibility of each option $o_i$ given the question $q$ with the information from both text $c$ and graph $\mathcal{G}_i$. Specifically, we concatenate the final representation of the global node with the representation of $[CLS]$ token from the pre-trained model and feed it into a classification layer.

\subsection{Neural-Symbolic Model: NSAR}
Although neural models are capable of well capturing language semantics, they disregard the interpretability of predictions.
To reconcile the robust learning in neural models and the discrete reasoning of symbolic methods, we design a neural-symbolic model to solve the AR task.
We propose an approach named NSAR (Neural-Symbolic model for Analytical Reasoning), which extracts the arguments and parses the constraints into compositional programs in a neural manner, and executes the programs to derive the answer with a symbolic inference engine. 

As the symbolic executor is non-differential, policy-gradient methods are usually employed to train the model~\cite{liang2016neural, bunel2018leveraging}. However, analytical reasoning questions can only be answered after examining the satisfiability of all the constraints, which leads to extremely sparse rewards and makes the model hard to be optimized. Thus, we manually annotate programs for supervision learning.

\subsubsection{Data Annotation}
We first extend the constraint function set defined in ARM \cite{zhong2021arlsat} to improve the scalability (e.g., \emph{FirstPos} and \emph{LastPos} are too specific).
It is composed of common logical functions (\emph{AND, OR, IF, NOT, etc.}) and operator functions including arithmetics ($+,-,=,>,<, etc.$), sorting (\emph{ARGMAX, MAX, etc.}), assignment (\emph{VALUE}), selection (\emph{SELECT}) and counting (\emph{COUNT}) operations.

Before program annotation, we need to annotate the participant and position sets of each problem.
Then for each constraint sentence in the context and the (question, option) pair, we annotate a program following the above definition, which is a composition of functions over the involved participants and positions. For example, we annotate a program as ``\emph{VALUE(roadster) $>$ VALUE(van) AND VALUE(roadster) $<$ VALUE(hatchback)}'' for the constraint ``\emph{The roadster is serviced later in the week than the van and earlier in the week than the hatchback}''. The statistics and examples of our data annotation are listed in Appendix A.

\subsubsection{Neural Parser}
In the neural perception module, we need to extract participants and positions from the context and interpret constraints into programs.

\paragraph{Participant/Position Extractor} We separately take participant extraction and position extraction as two sequence tagging tasks. For each task, a pre-trained model with a linear token classification head on top is employed for predicting which tokens are parts of participants or positions~\cite{liu2019roberta}.

\paragraph{Program Parser}
In order to model the relevance and dependency relationship between different constraints for program parsing, we take the whole context as an input and generate a sequence of compositional programs. The concatenation of the question and each option is also fed as an input to generate a program. Here we adopt a pre-trained encoder-decoder model for program generation~\cite{chen2019neural, raffel2019exploring}. Besides, to alleviate generating irrelevant or incorrect content in programs, we dynamically limit the vocabulary for each input to tokens of the input and all constraint functions.


\subsubsection{Symbolic Executor}
After obtaining programs from the context, question and each option, we need to execute these programs and present a score for each option. To do this, we inherit the executor from \cref{Program_exe} and \cref{option_select}.
We first execute programs from the context to obtain legitimate assignments. 
Then we execute the program of each (question, option) pair on these assignments and calculate the ratio of assignments satisfying the option as the score. We finally choose the option with the highest score as the answer.


\subsection{Results and Analysis}
\subsubsection{Overall Comparison}
We compare the performance of the above three methods with a baseline pre-trained model and human performance~\cite{zhong2021arlsat} on the validation and test sets. The pre-trained baseline model, CGAR and the extractor in NSAR all take RoBERTa-large~\cite{liu2019roberta} as the backbone while the parser in NSAR employs a T5-based model~\cite{raffel2019exploring}.
\begin{table}[!th]
	\centering
	\begin{tabular}{c|cc}
    \toprule
    Methods & Val (\%) & Test (\%) \\
	\midrule
	\emph{Human Performance} & - & 59.7 \\
	\midrule
    \emph{Random Guess} & 20.0 & 20.0 \\
	\midrule 
    \emph{ARM} & \bf 34.2 & \bf 30.9 \\
    \emph{RoBERTa} & 24.2 & 23.1 \\
    \emph{CGAR} & 27.7 & 24.9 \\
    \emph{NSAR} & 24.2 & 24.8 \\
	\bottomrule
 	\end{tabular}
	\caption{The answer prediction accuracy (\%) of different methods on AR-LSAT.}
	\label{table_result_AR_1}
\end{table}

The results are shown in Table~\ref{table_result_AR_1}. 
We observe that the symbolic system \emph{ARM} outperforms \emph{RoBERTa} and \emph{CGAR}. It shows that the complex analytical reasoning process is difficult to be completely parameterized by neural models, while \emph{ARM} works better by customizing complicated rules targeted at the AR-LSAT dataset and performing a deterministic deduction. Meanwhile, such difficulty for neural learning is exacerbated by limited data. 
From the improvement of \emph{CGAR} compared to \emph{RoBERTa}, multiple message-passing iterations over the constraint graph are proved to partly work on modeling the analytical reasoning process. 
However, \emph{NSAR} outperforms \emph{RoBERTa} while performing worse than both \emph{CGAR} and \emph{ARM}, because there is no large-scale supervised data. We annotate a small number of instances for extracting participant/position and parsing program, which results in interpreting imperfect programs and further influences the execution performance. We deem that annotating more data or heuristically augmenting data will boost the performance of \emph{NSAR}. 
Although we have achieved an improvement over the baseline model, our systems are still far from equivalent to human performance, leaving significant opportunities for further exploration of this highly challenging task.  



\subsubsection{Detailed Analysis on NSAR}
We first analyze the performance of participant extraction and position extraction in Table \ref{table_analysis_AR_extraction}.
The \emph{NSAR} performs relatively worse than \emph{ARM} in both validation and test sets, indicating that it is more appropriate to apply a rule-based extraction method to cover low-resource but diverse datasets.
In addition, there is still a certain gap between the \emph{ARM} model and perfect extraction, showing that AR-LSAT is extremely challenging and needs a finer set of extraction rules.
\begin{table}[!th]
	\centering
	\begin{tabular}{ccccc}
    \toprule
    \multirow{2}{*}[-0.7ex]{\centering Methods} & \multicolumn{2}{c}{Val} & \multicolumn{2}{c}{Test}\\
    \cmidrule(lr){2-3} \cmidrule(lr){4-5}
     & Prec. & Recall & Prec. & Recall  \\
	\midrule
	\emph{ARM} & 96.2 & 92.9 & - & -  \\
    \emph{NSAR} & 87.5 & 86.0 & 85.8 & 87.9  \\
    \midrule \midrule
	\emph{ARM} & 84.4 & 85.8 & - & - \\
    \emph{NSAR} & 56.3 & 52.2 & 44.9 & 61.5  \\
	\bottomrule
 	\end{tabular}
	\caption{Performance (\%) of the participant extraction (top) and position extraction (bottom).}
	\label{table_analysis_AR_extraction}
\end{table}

\begin{table}[!th]
	\centering
	\begin{tabular}{ccc}
    \toprule
    Parser Settings & EM & Rouge-L \\
    \midrule
    \emph{Combination} & 0.0  & 60.9 \\
    \emph{Separation} & 16.2 & 68.4 \\
	\bottomrule
 	\end{tabular}
	\caption{Evaluation results (\%) of the program parser on AR-LSAT. EM means Exact Match score. 
	}
	\label{table_analysis_AR_program_parser}
\end{table}
To investigate the performance of our program parser, we evaluate the programs generated by combining all sentences in the context and the question-option pair~(as \emph{Combination} setting) and by separately feeding each sentence or the question-option pair~(as \emph{Separation} setting), respectively.
As shown in Table~\ref{table_analysis_AR_program_parser}, our generated programs achieve a high Rouge-L while the exact match score is low and even equals zero for the combination setting. Besides the limited training data, it is because that program parsing itself is a challenging task. Specifically, it is susceptible to generating invalid programs by a small midterm mistake and suffers from the intrinsic diversity of programs that a constraint description can be formulated as different programs. Moreover, we observe that the program generation for long context is much more difficult than shorter sentences. How to design a simple and generic program language is an essential direction for saving annotation resources and improving program generation quality.

\subsubsection{Error Analysis}
We further dive into the error cases within our methods and summarize the major reasons causing wrong predictions for AR problems. The first reason is that participants and positions sometimes fail to be correctly extracted, which fundamentally causes the misunderstanding of the problem settings and thus affects the answer prediction of the aforementioned three methods. 

Some other errors occur because the predefined program language is not designed perfectly to cover some constraint descriptions with complex semantics. For example, ``two consecutive'' in the constraint ``\emph{No breed is featured on any two consecutive days.}'' is difficult to be formulated. The obstacle performance of the program parsing method also hinders the correct analytical deduction.

Although the textual descriptions in AR-LSAT are straightforward, basic commonsense knowledge is also needed to fully understand problem descriptions. For example, the system should realize that ``\emph{9:00 A.M.}'' and ``\emph{2:00 P.M.}'' are respectively in the morning and afternoon when it needs to arrange the schedule to satisfy  ``\emph{Some participants should be scheduled in the morning.}''.

\begin{figure*}[!th]
\centering
\includegraphics[width=1.95\columnwidth]{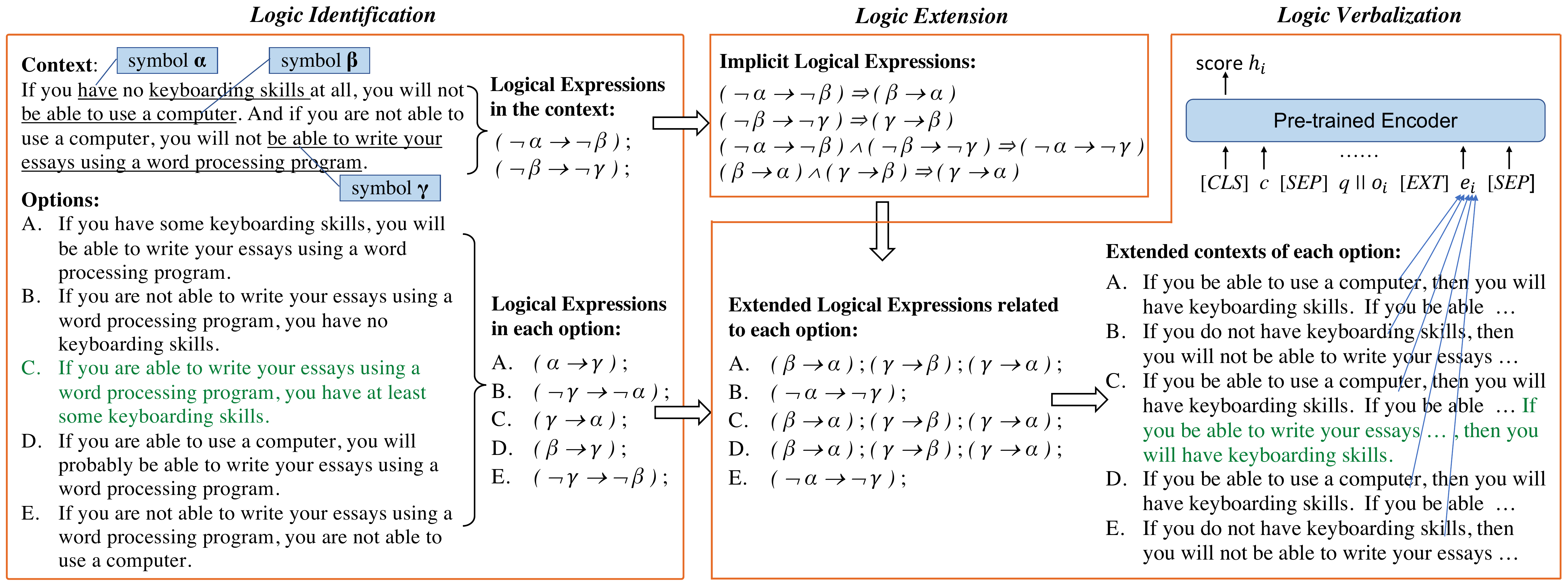}
\caption{\label{figure_framework} The overall architecture of the logic-driven context extension framework for LR. $c$, $q$, $o_i$ and $e_i$ are the context, question, $i$-th option and the extended context for $i$-th option, respectively. The texts in \textcolor[RGB]{74,163,87}{green} mean that the option $B$ is matched against its extended context which has the highest score.}
\end{figure*}
\subsubsection{Discussion}

We make different attempts on AR-LSAT, including neural, symbolic and neural-symbolic models, and observe that there is a certain gap between human performance and all attempted methods. Our models achieve a nearly 31\% accuracy while the powerful pre-trained models obtain only random performance.
We therefore would like to highlight the challenges of analytical reasoning and shed a light on the potential directions. 

In order to solve the AR problem, the main step is to exactly understand and abstract the textual constraints to machine cognitive programs with minimum manual effort. Moreover, a question is usually answered through multiple constraints. The research of modeling multiple constraints and mitigating error accumulation can be further studied. As commonsense knowledge is required for better understanding constraint descriptions, how to inject external knowledge into the AR task is also a further research direction.
Although analytical reasoning assesses both deep analytical understanding and complex deduction reasoning capabilities, only few data is available for exploration.
How to develop solid complex reasoning ability and ease the dependency on the data size should be given top priority in future research. 
Data augmentation is yet another feasible direction.



\section{Logical Reasoning}
\label{logical_reasoning}
\subsection{Challenges in Logical Reasoning}
Logical reasoning requires understanding a given text at a logical level and performing logical inference to deduce implications from asserted ones. It is a challenging task widely studied in recent years and several logical reasoning benchmarks have been introduced, such as ReClor~\cite{yu2020reclor} and LogiQA~\cite{liu2020logiqa}. 
Previous work usually treats the task as a traditional reading comprehension problem, and utilizes large-scale pre-trained language models~\cite{devlin2019bert, liu2019roberta, lan2019albert} or graph neural networks ~\cite{huang2021dagn} to match the context with candidate options.
Although promising results have been achieved, they mainly rely on word-level semantics without capturing symbolic logic.

In order to solve questions in LR-LSAT, as the example in Figure~\ref{figure_framework}, the reasoning system needs to extract the critical constituents from the context as logical symbols like ``$\alpha$: \emph{have keyboarding skills}'', ``$\beta$: \emph{be able to use a compute}'', ``$\gamma$: \emph{be able to write your essay using a word processing program}'' and identify the logical relationships between them to constituent existing logical expressions, like $(\neg \alpha \rightarrow \neg \beta)$ and $(\neg \beta \rightarrow \neg \gamma)$.
Then according to equivalence laws, it performs logical inference to extend the logical expressions that are not explicitly mentioned in the context.
Comparing the extended expressions with the expressions of candidate options, it selects the most similar option as the answer.

\subsection{LReasoner Model}
We propose a logic-driven reasoner (LReasoner) model for logical reasoning problems~\cite{wang2021logicdriven} from a neural-symbolic perspective.
It utilizes a logic-driven context extension framework to integrate the above reasoning process. Besides, a logic-driven data augmentation algorithm is also introduced, to construct literally similar but logically different samples and utilize contrastive learning to encourage our model better capture logical information. 

\subsubsection{Logic-Driven Context Extension}
The overall logic-driven context extension framework is illustrated in Figure~\ref{figure_framework}. It first identifies the logical symbols and expressions explicitly mentioned in the context as the elementary components for reasoning (Logic Identification). Then it performs logical inference following equivalence laws to extend the implicit logical expressions (Logic Extension). Finally, it verbalizes the extended logical expressions related to each option as an extended context and utilizes it into the pre-trained model to match the answer (Logic Verbalization).
\paragraph{Logic Identification} It employs a constituency parser \cite{joshi2018extending} to extract constituents including noun phrases and gerundial phrases from the context as basic logical symbols. Then the logical symbols in each sentence are combined by logical negative and conditional connectives $\{\neg, \rightarrow\}$ to constitute the logical expression as a follow-up.
If any negative word among \{``\emph{not}'', ``\emph{n't}'', ``\emph{unable}'', ``\emph{no}'', ``\emph{few}'', ``\emph{little}'', ``\emph{neither}'', ``\emph{none of}''\} is in or immediately before
a logical symbol $\alpha$, we add the negation connective $\neg$ before $\alpha$ as $\neg \alpha$.
If there is a conditional relationship between two logical symbols $\alpha$ and $\beta$ in a sentence, such as \emph{``if $\alpha$, then $\beta$'', ``$\alpha$ thus $\beta$'', ``$\beta$ due to $\alpha$'' and ``$\neg \beta$ unless $\alpha$''}, we can construct the corresponding logical expression as $(\alpha \rightarrow \beta)$. 
As illustrated in Figure~\ref{figure_framework}, it extract three logical symbols $\{\alpha, \beta, \gamma\}$ and identify two existing logical expressions as $(\neg \alpha \rightarrow \neg \beta)$ and $(\neg \beta \rightarrow \neg \gamma)$.
\paragraph{Logic Extension} 
There still exists some other implicit logical expressions which can be deduced from asserted ones. Therefore, it integrates the identified logical expressions in all sentences of the context, and performs logical inference over them to further extend the implicit expressions according to logical equivalence laws, including \emph{contraposition}~\cite{russel2013artificial} and \emph{transitive law}~\cite{zhao1997static}: 
\begin{align}
    &\text{Contraposition}: \ (\alpha \rightarrow \beta) \implies (\neg \beta \rightarrow \neg \alpha)  \label{Contraposition}\\
    &\text{Transitive Law}: \ (\alpha \rightarrow \beta) \ \land \ (\beta \rightarrow \gamma) \implies (\alpha \rightarrow \gamma)  \label{Transitive}
\end{align}
As shown in Figure~\ref{figure_framework}, it implies a set of extended logical expressions as $\mathcal{S}_E=\{(\beta \rightarrow \alpha), (\gamma \rightarrow \beta), (\neg \alpha \rightarrow \neg \gamma), (\gamma \rightarrow \alpha)\}$.
\paragraph{Logic Verbalization} 
Considering that symbolic logic is more difficult to encode, it uses the pre-trained model as the backbone of our framework. It verbalizes extended logical expressions into natural language and feeds them as an extended context into the pre-trained model. Specifically, it selects the related extended expressions for each option that have the same logical symbols as the option and transforms such expressions into natural language by filling them into a template. $(\neg \alpha \rightarrow \neg \gamma)$ can be verbalized as ``\emph{If do not $\alpha$, then will not $\gamma$}''. 
It takes such a sentence as an extended context for each option and feeds $[CLS] \ c \ [SEP] \ q \ || \ o \ [EXT] \ e \ [SEP]$ into the pre-trained model to get each option’s score.

\subsubsection{Logic-Driven Data Augmentation}
In order to make our model better capture logical information from the context, we also introduce a logic-driven data augmentation algorithm. It utilizes logical expressions to augment challenging samples with literally similar but logically different contexts.
Taking the original context to construct the positive sample, it constructs \emph{logical negative samples} by modifying the existing logical expressions in the context and verbalizing the modified expressions into a negative context. The modification operations include randomly deleting a logical expression, reversing the conditional order of a logical expression and negating a logical symbol in a logical expression.

It then adopts contrastive learning~\cite{chen2020simple} and trains our model to select the correct context supporting the answer to encourage the model to put more focus on logical information, especially logical negative and conditional relationships. As a result, the model is trained in a combination of answer prediction loss and context classification loss.

\subsection{Results and Analysis}
\subsubsection{Overall Comparison}
We compare our LReasoner model with several baseline models for logical reasoning, and the comparison results are shown in Table~\ref{table_result_LR_1}. We can see that \emph{$\text{LReasoner}_{\text{ALBERT}}$} achieves a great performance, outperforming all baseline models by a considerable margin on both validation and test sets. This indicates the effectiveness of our logic-driven system for logical reasoning. And \emph{$\text{LReasoner}_{\text{RoBERTa}}$} and \emph{$\text{LReasoner}_{\text{ALBERT}}$} both perform better than the corresponding baseline models \emph{RoBERTa} and \emph{ALBERT}. It demonstrates that \emph{LReasoner} is robust to be effective for logical reasoning on top of different pre-trained models.
\begin{table}[!th]
	\centering
	\begin{tabular}{ccc}
    \toprule
    Methods & Val (\%) & Test (\%) \\
	\midrule
    \emph{Random Guess} & 20.0 & 20.0 \\
	\emph{BERT} & 39.3 & 39.4 \\
    \emph{RoBERTa} & 49.2 & 49.6 \\
    \emph{ALBERT} & 57.9 & 57.8 \\
    $\text{\emph{LReasoner}}_{\text{\emph{RoBERTa}}}$ & 54.0 & 53.3 \\
    $\text{\emph{LReasoner}}_{\text{\emph{ALBERT}}}$ & \bf 65.0 & \bf 63.5 \\
    \midrule
    \textbf{Ablation study} \\
    \midrule
    $\text{\emph{LReasoner}}_{\text{\emph{ALBERT}}}$ (w/o \emph{CE}) & 63.4 & 61.6 \\
    $\text{\emph{LReasoner}}_{\text{\emph{ALBERT}}}$ (w/o \emph{DA}) & 61.7 & 60.0 \\
	\bottomrule
 	\end{tabular}
	\caption{The answer prediction accuracy (\%) of different methods on LR-LSAT. \emph{CE} and \emph{DA} are our logic-driven context extension framework and data augmentation algorithm.}
	\label{table_result_LR_1}
\end{table}

We also conduct ablation study which takes \emph{ALBERT} as our backbone model. \emph{$\text{LReasoner}_{\text{ALBERT}}$ (w/o CE)} and \emph{$\text{LReasoner}_{\text{ALBERT}}$ (w/o DA)} both outperform the baseline model \emph{ALBERT} and perform worse than our final system \emph{$\text{LReasoner}_{\text{ALBERT}}$}. It demonstrates that both logic-driven context extension framework and logic-driven data augmentation algorithm are beneficial for logical reasoning.


\subsubsection{Analysis of Logic Identification}
To investigate the performance of the heuristic logic identification
method, we randomly sample 50 instances and manually annotate the logical symbols and expressions as labels. We report the recall score of logical symbol identification and logical expression identification as 65.9\% and 48.9\%, respectively. We can see that our generic logic extraction method which operates in an unsupervised manner achieves relatively reliable performance. How to design an unsupervised and generic logic extraction method is essential to be studied to further enhance the performance of the overall system.

\subsubsection{Performance on ReClor \& LogiQA}
We also conduct experiments on two public logical reasoning datasets, ReClor~\cite{yu2020reclor} and LogiQA~\cite{liu2020logiqa}, to investigate the robustness of our logic-driven reasoner. ReClor dataset is proposed from GMAT and LSAT tests while LogiQA is collected from the National Civil Servants Examination of China, and each question is provided with a context and four answer options. As shown in Table~\ref{table_result_reclor}, our system is robust to be effective on both ReClor and LogiQA, and even surpasses the human performance of ReClor.

\begin{table}[ht]
\begin{center}
\begin{tabular}{ccccc}
\toprule
\multirow{2}{*}[-0.7ex]{\centering Model} &\multicolumn{2}{c}{ReClor} &\multicolumn{2}{c}{LogiQA}\\
\cmidrule(lr){2-3}\cmidrule(lr){4-5}
 & Val & Test & Val & Test\\
\midrule
\emph{RoBERTa}~\cite{liu2019roberta} & 62.6 & 55.6 & 35.9 & 35.3 \\
\emph{ALBERT}~\cite{lan2019albert} & 70.2 & 66.5 & 38.9 & 37.6 \\
\emph{DAGN}~\cite{huang2021dagn} & 65.8 & 58.3 & 36.9 & 39.3\\
$\text{\emph{LReasoner}}_{\text{\emph{RoBERTa}}}$ & 66.2 & 62.4 & 38.1 & 40.6 \\
$\text{\emph{LReasoner}}_{\text{\emph{ALBERT}}}$ & \bf 73.2 & \bf 70.7 & \bf 41.6 & \bf 41.2 \\
\midrule
\emph{Human Performance} & - & 63.0 & - & 86.0 \\
\bottomrule
\end{tabular}
\end{center}
\caption{\label{table_result_reclor} Experimental results (accuracy $\%$) of different models on ReClor and LogiQA.}
\end{table}

\subsubsection{Error Analysis}
Although our LReasoner system achieves outstanding performance, there still exist some instances that cannot be solved. Similar to the logical reasoning dataset ReClor~\cite{yu2020reclor}, LSAT also integrates various types of logical reasoning skills, including ``\emph{Necessary Assumptions}'', ``\emph{Sufficient Assumptions}'', ``\emph{Implication}'', ``\emph{Most Strongly Supported}'', ``\emph{Strength}'', ``\emph{Weaken}'', ``\emph{Match flaws}'', etc. We thus investigate the detailed performance with respect to different logical reasoning types to analyze which type of questions tend to be more challenging.

Among nearly 17 categories, our model performs relatively poorly on \emph{Match flaws} and \emph{Weaken} with an almost 60\% accuracy while accuracies of other types are higher than 70\%. \emph{Weaken} aims to find the statement that \textbf{weaken} the argument. \emph{Match flaws} is even more challenging as it requires analysis of the flaw that \textbf{conflicts} with the complete logical chain illustrated in the context, and find an option exhibiting the same flaw. Our system that first extracts logical expressions and then implies the implicit logical expressions is not suitable for abstracting the complete logical chain for flaw matching and modeling the different degrees of a logical statement to identify weaken statements.

\subsubsection{Discussion}
Our logic-driven system is able to model the discrete logical inference process explicitly and achieves an outstanding performance on LR-LSAT. Our system even surpasses human performance on another logical reasoning dataset, ReClor.
However, some challenges are still there. The first is that current methods treat all types of logical reasoning skills alike and do not dive into the difference between reasoning types. How to deal with different logical reasoning types is a further research direction. 
Moreover, the main challenge for logical reasoning is to point out the logical structures among the context. Specifically, automatically extracting the logical elementary units and identifying the logical relationships between units in an unsupervised manner is essential to be explored and improved. 
Further research can also focus on directly encoding the symbolic logical structure rather than verbalizing them into natural language for utilization. Concretely, we can separately model each logical connectives in different networks, and take the involved logical symbols as the network inputs to learn the logical expressions.

\section{Reading Comprehension}
\label{reading_comprehension}

\subsection{RC-LSAT Challenges}
\label{rc_challenge}
Apart from analytical reasoning and logical reasoning, LSAT also involves reading comprehension, which is a fundamental ability of human intelligence. It requires understanding long-form, complex passages and distinguishing what is the correct statement by synthesis, comparison, and application of principles. Many widely studied datasets have been developed for reading comprehension, such as SQuAD~\cite{rajpurkar2016squad}, MCTest~\cite{richardson2013mctest} and RACE~\cite{conf/emnlp/LaiXLYH17}.
Recent pre-trained language models~\cite{liu2019roberta, lan2019albert} and diverse attention modules~\cite{zhang2020retrospective, zhu2020duma} achieve state-of-the-art performance on them, and even exceed human performance.
We compare RC-LSAT with several similar multiple-choice reading comprehension datasets including MCTest, RACE, and COSMOS QA~\cite{huang2019cosmos} to investigate its challenges.
The overall statistics are presented in Table \ref{tab: compare_statistics}. 
\begin{table}[ht]
\centering
\begin{tabular}{l c c c c}
\toprule
Dataset & MCTest & RACE & COSMOS QA & RC-LSAT \\
\midrule
\# of contexts & 660 & 27,933  & 21,866 & 360 \\
\# of questions & 2,640 & 97,687 & 35,588 & 2,419 \\
\midrule
Context Len & 210.1 & 321.9 & 70.3 & 511.5 \\
Question Len & 7.8 & 10.0 & 10.6 & 20.5 \\
Option Len & 3.4 & 5.3 & 8.1 & 16.8 \\
\bottomrule
\end{tabular}
\caption{Statistics of RACE, COSMOS QA, and RC-LSAT.}
\label{tab: compare_statistics} 
\end{table}


We can see that the context, question and option of RC-LSAT are more complicated than other datasets with longer sequences, which are also more difficult to be comprehended.
Take a look at the following example question ``\emph{Which one of the following most accurately and completely describes the function of the second paragraph of the passage?}'' and the corresponding answer ``\emph{explains the ramifications of the strict constructionists claims and helps clarify the relevance of evidence offered in subsequent paragraphs}''. Not only text understanding and function abstraction, but also positional information modeling, such as ``\emph{the second paragraph of the passage}'', should be both incorporated for answer prediction. 
Besides, RC-LSAT is of relatively limited data size.
The above-mentioned challenges make models stuck in more complex reading comprehension with few data.


\subsection{Position-aware DUMA Model}
Considering the state-of-the-art performance achieved by DUMA model~\cite{zhu2020duma} on most similar examination-based reading comprehension datasets RACE involving reasoning, we also employ DUMA for RC-LSAT. Compared to the baseline Model introduced in Sec~\ref{sec:base_model}, it additionally employs a Dual Multi-head Co-Attention module between the pre-trained encoder model and the classification layer. It simulates human transposition thinking patterns to further capture relationships of key information from the passage, question and answer options. 
In Dual Multi-head Co-Attention module, it first separates the output representation of the encoder to obtain 
$E^P=[e^p_1,e^p_2,...,e^p_{l_p}]$ and 
$E^{QA}=[e^{qa}_1,e^{qa}_2,...,e^{qa}_{l_{qa}}]$,
where $e^p_i$, $e^{qa}_j$ denote the $i$-th and $j$-th token representation of passage and question-answer respectively and $l_p$, $l_{qa}$ are the corresponding lengths.
It then calculates question-answer-aware passage representation $E^{P(QA)}$ and passage-aware question-answer representation $E^{QA(P)}$ in a bi-directional way. It aggregates the key information from $E^{QA(P)}$ and $E^{P(QA)}$ as $O_i$ by mean pooling and concatenation operations and utilizes it for answer classification instead of $[CLS]$ representation.
\begin{equation}
\begin{split}
    E^{QA(P)} &=MultiHeadAttn(E^P, E^{QA}, E^{QA}), \\
    E^{P(QA)} &=MultiHeadAttn(E^{QA}, E^P, E^P), \\
    O_i = &[Mean(E^{QA(P)}); Mean(E^{P(QA)})], \\
\end{split}
\end{equation}

Note that questions of RC-LSAT usually involve positional information indicators, like ``\emph{line 3-5}'' and ``\emph{second paragraph}'', we label the context with position marks, e.g., ``$\langle$line3$\rangle$... $\langle$/line3$\rangle$'' for line 3 and ``$\langle$P2$\rangle$...
$\langle$/P2$\rangle$'' for paragraph 2. We feed labeled context as input and implement a position-aware DUMA model.

\subsection{Results and Analysis}
\subsubsection{Overall Comparison}

We compare our position-aware \emph{DUMA} (\emph{P-DUMA}) model with several baseline models and \emph{DUMA} model and the comparison result is shown in Table~\ref{tab: mrc_result}. We can see that our \emph{P-DUMA} model achieves outstanding performance on both validation and test sets. Position marks are observed to help improve performance on top of both \emph{ALBERT} and \emph{DUMA}, which shows the effectiveness of our proposed position mark for RC-LSAT.
\begin{table}[ht]
\centering
\begin{tabular}{l c c}
\toprule
Model & Val (\%) & Test (\%) \\
\midrule
\emph{RoBERTa} & 52.0  & 44.0 \\
\emph{ALBERT}  & 51.9  & 48.7 \\
\emph{DUMA}  & \bf 60.0  & 52.0  \\
\emph{P-ALBERT} & 58.5 & 55.4  \\
\emph{P-DUMA} &  57.4  & \bf 56.1 \\
\bottomrule
\end{tabular}
\caption{Overall performance (\%) on RC-LSAT. \emph{P-ALBERT} and \emph{P-DUMA} mean position-aware ALBERT and our position-aware DUMA model, respectively.}
\label{tab: mrc_result} 
\end{table}


\subsubsection{Transfer Learning}
Previous work~\cite{min2017question, jin2020mmm} has shown the effectiveness of pre-training on similar datasets then fine-tuning on the target dataset for transfer learning, which can partly relieve the pressure of data sparsity. Due to limited samples in the RC-LSAT, we adopt the most similar RACE as the source dataset and conduct a set of transfer learning experiments. As shown in Table~\ref{tab: mrc_transfer_result}, a significant improvement is obtained after transfer learning for all models, which further validates the potential of transfer learning for reading comprehension. However, positional information has no effect in transfer learning settings compared to position-unaware \emph{ALBERT} and \emph{DUMA}. The reason may be that the gap between RC-LSAT and RACE becomes larger after using position marks, as no position label is provided in RACE. 

\begin{table}[ht]
\centering
\begin{tabular}{l c c}
\toprule
Model & Val (\%) & Test (\%) \\
\midrule
$\emph{ALBERT}_{\text{RACE}}$  & \bf 73.7  & 67.7 \\
$\emph{DUMA}_{\text{RACE}}$  & 71.9  & \bf 69.1  \\
$\emph{P-ALBERT}_{\text{RACE}}$ & 71.5 & 68.4  \\
$\emph{P-DUMA}_{\text{RACE}}$ &  68.5  & 63.9 \\
\bottomrule
\end{tabular}
\caption{Transfer learning results of models first trained on RACE and then fine-tuned on RC-LSAT.}
\label{tab: mrc_transfer_result} 
\end{table}


\subsubsection{Error Analysis}
Although outstanding performance has been achieved, there still exist some challenging instances. We illustrate the major error types of RC-LSAT as follows.

The first type of error is caused by the reason that the context is too long and some essential information for answer prediction is truncated and cannot be effectively encoded. 
The second error type is that some comparative questions require comparative learning ability, which still has not been considered.
Take the question ``\emph{The authors of the passages would be most likely to disagree over whether?}'' as an example, the model not only needs to understand the view of each author, but also aims to comparatively infer the point that authors would disagree with.
In addition, the lack of commonsense knowledge is another key problem for the wrong prediction. For example, unaware that 1934 and 1939 are within the 1930s, the model would fail to answer the question ``\emph{According to the passage, which one of the following was true of the physics community during the 1930s?}''.

\subsubsection{Discussion}
Although previous work~\cite{lan2019albert, zhu2020duma} and our models have shown promising results on RC-LSAT, we still have a way to go to improve performance.  
As the truncation mechanism of pre-trained models threatens modeling long sequence and the sparse self-attention mechanism~\cite{beltagy2020longformer} sometimes fails to be effective, designing an appropriate hierarchical encoder for long-form sequence encoding is emerging as an area for further exploration. 
Besides, current methods rarely specialize in answering comparative questions, how to comparatively reading multiple passages and diving into the differences also requires potential research.

A more challenging direction is to retrieve relevant commonsense or passages from open source and incorporate them as evidence for answer prediction.
Previous work~\cite{conf/acl/SunLLJ18,conf/aaai/PangLGXSC19} can only achieve poor performance on this topic due to intrinsic noise in retrieved passages. Further research can focus on improving evidence retrieval and predicting answers from noised evidence. 


\section{Further Discussion}
\label{overall_discussion}
After making attempts on three tasks of the LSAT, we have achieved some progress towards complex reasoning. We evaluate our whole system on the LSAT tests and raise some positive findings. We further discuss the existing challenges of complex reasoning and shed a light on the future research directions.

\subsection{Overall Performance of LSAT Tests}
To give an intuitive overview of our machine intelligence in LSAT tests, we convert our raw accuracy scores of three tasks to an LSAT scale, which ranges from 120 to 180~\cite{lsat2020conversion}. We also integrate the accuracies of AR, LR and RC, to calculate an overall score by weighted averaging with the original proportion of $1:2:1$. We compare our scaled scores with the median score of candidates taking LSAT exams during $2019\sim2020$~\cite{lsac2020scores}. We further demonstrate which level of schools our systems could be admitted to according to the law school rankings displayed by Internet Legal Research Group~\cite{lsat2020ranking}.
\begin{table}[ht]
\centering
\begin{tabular}{l c c}
\toprule
 & Scaled (Raw) Score & School Ranking \\
\midrule
Overall system & 151 (56.8\%) & Top 104 \\
AR system & 135 (30.9\%) & $>200$ \\
LR system & 155 (63.5\%) & Top 58\\
RC system & 158 (69.1\%) & Top 30\\
\midrule
Candidates & 152 (58.0\%) & Top 94 \\
\bottomrule
\end{tabular}
\caption{Comparison results of our systems with human candidates. 
\emph{$>200$} means that the ranking of AR system is beyond the top 200 which is not displayed by Internet Legal Research Group~\cite{lsat2020ranking}.}
\label{tab: lsat_compare_statistics} 
\end{table}

As shown in Table~\ref{tab: lsat_compare_statistics}, we have several positive findings. Our overall system achieves impressive performance on the standard LSAT tests designed to examine the reasoning ability of prospective law school candidates, and performs comparably with the median candidate scores, which indicates the potential of machine complex reasoning.
Our systems for RC and LR even have a chance to be accepted by the top 30 and 58 law schools, respectively, which demonstrates that the combination of the powerful pre-trained models and task-specific reasoning modules is effective in performing complex reasoning. 
Concretely, our logic-driven system endows itself an excellent logical reasoning ability by performing explicit logical inference while the DUMA model aware of position information with transfer learning possesses fundamental complex reading comprehension capability. Although AR is an extremely challenging task requiring a more comprehensive understanding of all context pieces to build the whole reasoning chain and only few data is available, our symbolic ARM system is still able to get into a law school, which shows that symbolic knowledge plus discrete interpretable reasoning steps is essential in solving analytical reasoning task.


\subsection{Challenges \& Future Directions}
Despite the positive achievement, some unsolved challenges remain in complex reasoning. We investigate the major challenges and the corresponding potential solutions as follows.
\subsubsection{Unsupervised Symbolic Knowledge Extraction} 
The automatic extraction of elementary symbolic units or expressions builds the foundation of complex reasoning tasks as it is required to fully understand complex scenarios, which further affects the performance of the overall reasoning system. For example, the extraction of mathematical expressions composed of quantities and arithmetic signs is essential for numerical reasoning, and logical reasoning is heavily reliant upon logical expression identification.
However, predefined rule patterns by domain experts or large-scale annotated data for symbolic knowledge extraction are expensive, which are impractical to be obtained for all the tasks. Therefore, the unsupervised extraction of symbolic knowledge is a major challenge in complex reasoning. 

To handle this challenge, we can begin by utilizing formal programming languages~\cite{rawlings1985reasoning, boulton1997tool} to design a generic and extensible extraction framework for universal symbolic knowledge, supplemented by a set of task-specific extension and modification operations. In this way, we can further modify and extend the generic framework, and obtain the specific extraction method for different complex reasoning tasks to automatically extract high-quality symbolic knowledge.

\subsubsection{Model Interpretability}
Interpretability is a significant characteristic of trustworthy and controllable reasoning systems, which makes the decision-making process of complex reasoning comprehensible and predicts more reliable results. Besides, the incorrect prediction can be traced to learn the cause and intervened to make a revised prediction. For instance, the explicit walk along the multi-hop relational paths makes the multi-hop reasoning process more interpretable to find the answer.
Although neural models achieve robust performance on complex reasoning, their prediction is always a black box for humans to understand. How to improve the interpretability of a neural system is essential to be studied.

We can design the neural model structure from the perspective of simulating human cognition and the reasoning process. It integrates multiple modules for different reasoning steps, and injects the intermediate results into the whole reasoning chain, which makes the neural model interpretable.

\subsubsection{Few-Shot Learning}
As the data targeted at complex reasoning with high quality and difficulty is rare and hard to be collected or annotated, the traditional data-driven learning methods heavily relying on a huge amount of training
instances may have poor performance on complex reasoning tasks. The analytical reasoning task is a typical example with few data. Therefore, a few-shot learning paradigm~\cite{fei2006one, xiong2018one, du2019cognitive} urgently needs to be explored for complex reasoning to improve the reasoning ability and the generalization capability of systems with only a few training instances. 

Transfer learning will be a potential direction of few-shot learning for complex reasoning. It uses the pre-training models on related tasks as the starting point which can ease the sufferings of data sparsity, and transfer the reasoning ability learned from source task to target domain. We can also heuristically synthesize the data by modifying original data to augment the dataset of the current complex reasoning task.

\subsubsection{All-Sided Benchmark for Complex Reasoning}
Recent years have witnessed an increasing trend in natural language understanding towards complex reasoning, yet there is no existing integrated benchmark to the best of our knowledge, that comprehensively evaluates different types and domains of complex reasoning ability. It is worthwhile to build an all-sided benchmark dataset to promote research in complex reasoning.

First of all, the benchmark should cover the three reasoning capabilities involved in the LSAT. Specifically, reading comprehension establishes the foundation of complex reasoning to understand and summarize the semantics of substances and qualities. Logical reasoning extends complex reasoning ability with logical deduction over propositions while analytical reasoning simulates the human analytical thinking and problem-solving capacity. Then several widely studied complex reasoning tasks should be integrated, such as reasoning for commonsense knowledge~\cite{talmor2018commonsenseqa, speer2017conceptnet, sap2019atomic}, multi-hop relationships~\cite{welbl2018constructing, yang2018hotpotqa} and numerical calculation~\cite{dua2019drop, amini2019mathqa} described in~\cref{reasoning_taxonomy}. 

Some other rarely explored complex reasoning abilities also need to be considered. For example, we should dive into abductive reasoning~\cite{bhagavatula2019abductive} and counterfactual reasoning~\cite{starr2019counterfactuals, qin2019counterfactual} to research cause-and-effect complex reasoning.
The former aims to trace the most likely explanation to partial observations and the latter focus on the illation of how alternative events in the past result in different outcomes. Analogical reasoning~\cite{bartha2013analogy} that draws a comparison between different things and based on their similarities to infer their further shared properties is widely adopted in human daily thinking.
As no benchmark dataset so far has been proposed for mainly enhancing analogical inference, it also should be involved in this comprehensive benchmark.

\section{Conclusion}
\label{conclusion}
In this paper, we take a step towards complex reasoning research by studying three tasks involved in LSAT tests, namely, analytical reasoning, logical reasoning and reading comprehension. Inspired by the advantages and disadvantages of symbolic, neural, and neural-symbolic methods of complex reasoning, we propose a hybrid system for three tasks and achieve promising performance on the LSAT tests. In particular, our position-aware DUMA model with transfer learning for reading comprehension and logic-driven model for logical reasoning even have a chance to be recruited to the top 30 and top 58 law schools, respectively. It demonstrates our progress in modeling complex reasoning abilities, notably the fundamental reading comprehension and challenging logical reasoning capability.
Through a systematical study of the LSAT, we further overall discuss the unsolved challenges in complex reasoning and investigate the potential directions. In the future, how to extract symbolic knowledge in an unsupervised manner and improve the interpretability of a neural reasoning system are promising directions.
Few-shot complex reasoning and building a comprehensive complex reasoning benchmark are also worthy of exploration.


%

\appendices
\section{Data Annotation}
\begin{figure}[!th]
\centering
\includegraphics[width=0.97\columnwidth]{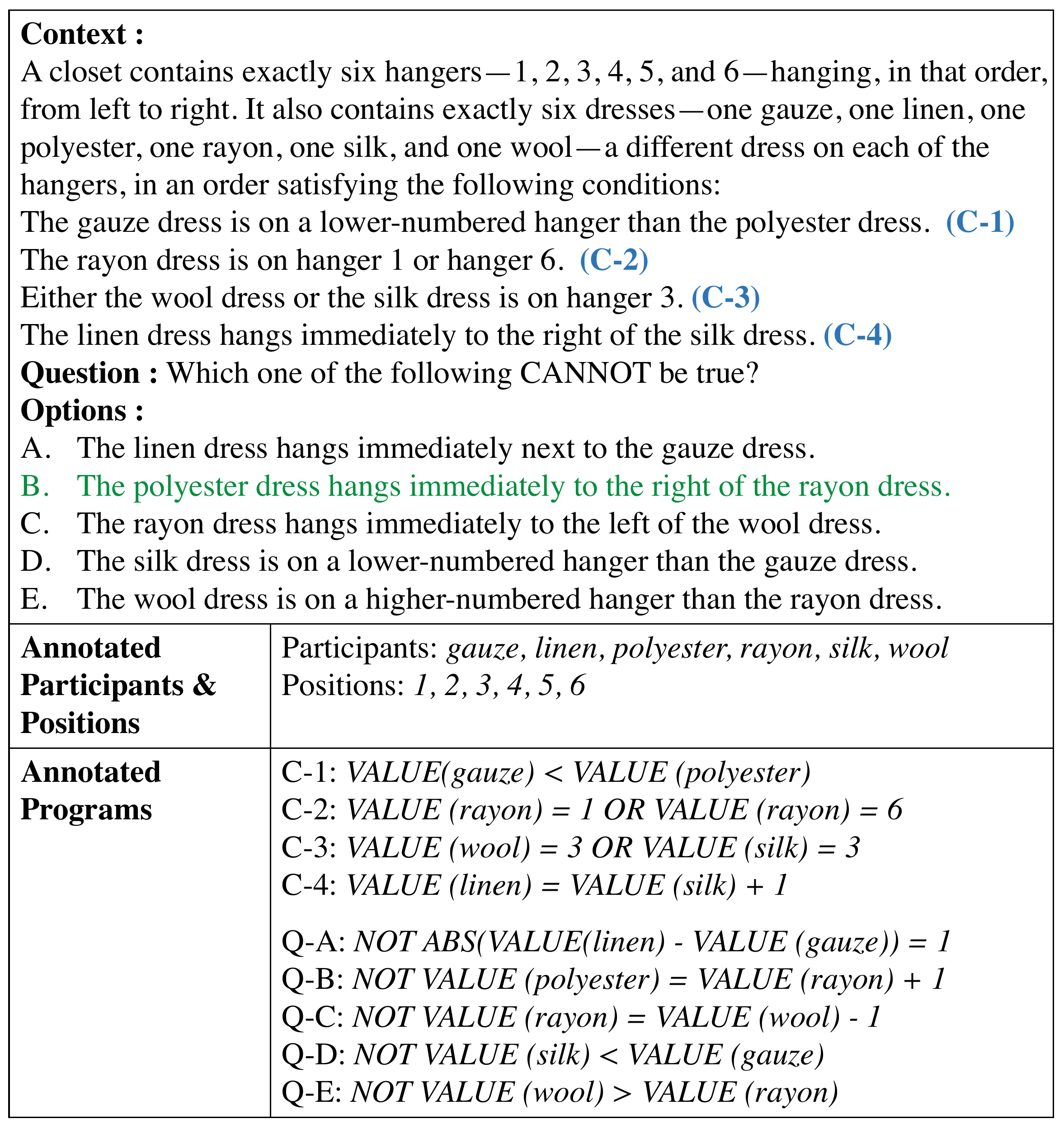}
\caption{\label{figure_annotation} An example of our annotation for an AR problem. C-$i$ means the $i$-th constraint while Q-$j$ denotes the pair of the question and option $j$.}
\end{figure}

We annotate a subset of programs to investigate the operation mechanism of our neural-symbolic method. 
An example of our annotation is shown in Figure~\ref{figure_annotation}. For each problem, we not only annotate the participants and positions, but also annotate the programs of all constraints in the context and question-option pairs.
We annotate a total of $100/40/40$ instances for participants and positions extraction in the training/validation/test set. 
As the AR task has a situational property that a group of questions share the same context but ask different aspects of information, we annotate programs for total 77 contexts and 428 questions, with $50/13/14$ contexts and $274/75/79$ questions in the train/validation/test set. 

\ifCLASSOPTIONcaptionsoff
  \newpage
\fi



\bibliographystyle{IEEEtranN}
\bibliography{taslp}
\end{document}